\documentclass[letterpaper]{article} 
\usepackage{flairs}
\usepackage{times}
\usepackage{helvet}
\usepackage{courier}
\usepackage{amsmath,amsfonts,mathtools}
\usepackage[hyphens]{url}  
\usepackage{graphicx} 
\usepackage{graphicx}  
\usepackage{paralist}
\usepackage{subcaption}
\usepackage{multirow}

\usepackage{booktabs}

\begin{document}

\title{Visual Episodic Memory-based Exploration}

\author{Jack Vice\\
Dept. of Computer Science\\
University of Central Florida\\
Orlando, USA\\
\And 
Natalie Ruiz-Sanchez\\
Dept. of ECE\\
University of Central Florida\\
Orlando, USA\\
\And 
Pamela K. Douglas\\
Dept. of Modeling and Simulation\\ 
University of Central Florida\\
Orlando, USA\\
\And 
Gita Sukthankar\\
Dept. of Computer Science\\
University of Central Florida\\
Orlando, USA\\
}

\maketitle

\begin{abstract}
In humans, intrinsic motivation is an important mechanism for open-ended cognitive development; in robots, it has been shown to be valuable for exploration.
An important aspect of human cognitive development is \textit{episodic memory} which enables both the recollection of events from the past and the projection of subjective future.  This paper explores the use of visual episodic memory as a source of intrinsic motivation for robotic exploration problems.  Using a convolutional recurrent neural network autoencoder, the agent learns an efficient representation for spatiotemporal features such that accurate sequence prediction can only happen once spatiotemporal features have been learned.  Structural similarity between ground truth and autoencoder generated images is used as an intrinsic motivation signal to guide exploration.  Our proposed episodic memory model also implicitly accounts for the agent's actions, motivating the robot to seek new interactive experiences rather than just areas that are visually dissimilar.  When guiding robotic exploration, our proposed method outperforms the Curiosity-driven Variational Autoencoder (CVAE) at finding dynamic anomalies.
\end{abstract}

\section{Introduction}
Although much animal behavior is driven by the immediate need to maintain homeostasis while escaping predators, humans possess the \textit{intrinsic motivation} ``to explore, manipulate or probe their environment, fostering curiosity and engagement in playful and new activities''~\cite{oudeyer2007}.  Intrinsic motivation is an important mechanism for promoting lifelong sensorimotor and cognitive development. For autonomous agents, it is particularly valuable for reinforcement learning problems with a sparse reward structure, and curiosity-driven learning has been shown to be effective even in the absence of extrinsic rewards~\cite{pathak2017curiosity}.  However, meaningfully representing intrinsic motivation remains an open ended research question.  Oudeyer and Kaplan~\shortcite{oudeyer2007} presented a large typology of computational approaches for quantifying intrinsic motivation, including  knowledge-based, learning progress, competence, and morphological models, yet many of these models remain as yet untested.  The usage of these techniques within robotics suffers from a chicken and egg problem: intrinsic motivation is critical for creating a complex cognitive reasoning system capable of lifelong, open-ended learning yet it is impossible to utilize sophisticated models of intrinsic motivation without a more complex cognitive architecture.   This paper proposes a stepping stone to this problem by imbuing the robot with visual episodic memory in the form of a convolutional recurrent neural network autoencoder.  Most mobile robots already possess sophisticated object detection systems grounded in deep learning, making it  feasible to augment existing perceptual systems with our proposed episodic memory architecture.

Unlike semantic memory which stores facts, ideas, and concepts, episodic memory centers around individual recollections.
Humans experience both the past and the future through episodic memory; it can be viewed as a form of ``mental time travel''~\cite{tulving2002}, enabling the recollection of past events and the projection
of subjective future~\cite{wheeler1997}. Human visual episodic memory is massive and complex; it has been noted that ``people remember a lot about the things that they see over a lifetime, and they remember with a level of precision that remains out of reach for artificial systems''~\cite{schurgin2018}.  Hence we believe that our visual episodic memory model provides a suitably rich substrate for calculating intrinsic motivation.

In the context of search and rescue as well as security robots, the ability to rapidly search an area for anomalous spatiotemporal features is important.  Example applications include searching for areas of a building that are structurally damaged, patrolling for intruders, and rescuing casualties.
Ramakrishnan et al.~\shortcite{ramakrishnan2021exploration} categorize approaches to embodied visual exploration as prioritizing \textit{novelty}~\cite{Bellemare2016unifying,Ostrovski2017count}, \textit{coverage}~\cite{chen2019learning}, \textit{curiosity}~\cite{oudeyer2007,pathak2017curiosity} or \textit{reconstruction}~\cite{ramakrishnan2018sidekick}.
Novelty seeks unvisited states, whereas coverage algorithms try to reveal unseen states in the environment. Curiosity-driven exploration seeks areas with high uncertainty, and reconstruction approaches try to visit states that may help predict other unseen states.  Our visual episodic memory model is a curiosity-based exploration approach since it prioritizes states that are poorly predicted by the convolutional recurrent network autoencoder.   
This paper makes the following research contributions:
\begin{compactenum}
\item introduces a new visual episodic memory architecture that uses a convolutional LSTM autoencoder to learn a compressed representation of the robot's visual experiences;
\item introduces the usage of multi-frame Structural Similarity Index Measure (SSIM) as a metric for episodic learning (a form of intrinsic motivation) to guide the robot to explore areas where the visual prediction is faulty;
\item demonstrates that our method outperforms both frontier-based exploration and  curiosity methods such as the Curiosity-driven Variational Autoencoder (CVAE) at finding dynamic anomalies in simulation; 
\item illustrates the usage of our convolutional LSTM autonencoder at reconstructing videos from a small mobile robot.
\end{compactenum}



\begin{figure}
\begin{center}
\includegraphics[scale=.35]{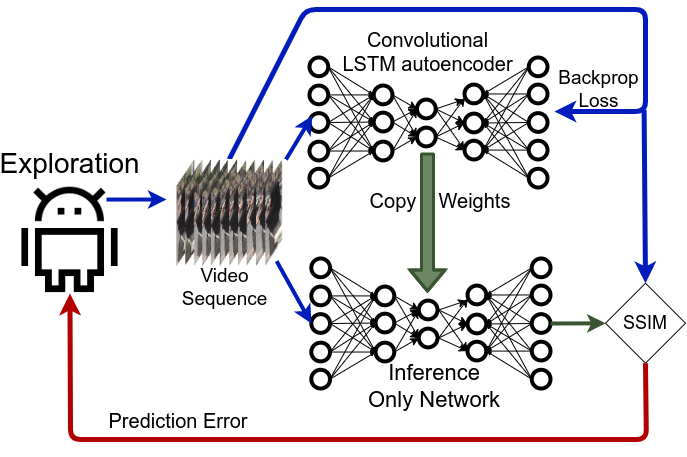}
\caption{Our architecture consists of a simulation environment, twin convolutional LSTM autoencoders and a frontier exploration based navigation stack.  The twin models run asynchronously with weights copied from the training model to the inference only model, enabling faster predictions for the mobile robot.}
\label{fig:sys_arch}
\end{center}
\end{figure}

\section{Related Work}
There is a rich history of research efforts dedicated to the problem of exploring static environments with a mobile robot~\cite{brooks1986robust,kuipers1991robot,giralt1990integrated}.  When mapping the environment is the main objective of exploration, frontier exploration~\cite{yamauchi1997frontier} can be used in combination with Simultaneous Localization and Mapping (SLAM)~\cite{durrant2006simultaneous}.
However, these techniques assume that the environment is static; a dynamic environment requires modeling distributions of relevant features over time as well as space.  Exploration algorithms for static environments simply determine ``Have I been here before?'', whereas we aim to answer the  question ``Have I interacted with the objects in this area before?''

In humans, this question is answered by consulting one's visual episodic memory. Episodic memory~\cite{tulving197212} is a type of autobiographical memory involving specific events, personal facts, and flashbulb memories~\cite{brown1977flashbulb}. Unlike semantic memory, the episodic memory system supports mentally replaying events and is important for achieving autonoetic consciousness of one's identity stretching from the past to the future~\cite{gardiner2008}. In humans, visual episodic memory can provide a perception-like experience in which one inhabits a virtual world recreated in memory~\cite{lin2018}. Schurgin and Flombaum~\shortcite{schurgin2018} investigated how changes in view orientation and the spacing of encounters affect visual episodic memory following repeated encounters with objects. Our paper proposes implementing a robot's visual episodic memory using a convolutional LSTM autoencoder which we believe would accurately model some of the findings described by Schurgin and Flombaum. 

Chong and Tay~\shortcite{chong2017abnormal} introduced a spatiotemporal autoencoder for the purpose of detecting abnormal events in videos. Their semi-supervised training method uses convolutional layers to learn the spatial encoding and LSTM layers for temporal encoding.  After training on baseline video sequences of dynamic scenes, the convolution LSTM autoencoder is able to detect anomalous events in video data. This work builds on \citeauthor{hasan2016learning} \shortcite{hasan2016learning} in which  a convolutional autoencoder was used to predict regularity in the videos.  Rothfuss et al.~\shortcite{rothfuss2018} proposed a similar convolutional LSTM architecture trained with two decoders (present and future) for modeling episodic memory in robots; however they applied it to a case based retrieval system for identifying previously experienced problems to solve new situations. In contrast, our aim is to extract an intrinsic motivation signal from visual episodic memory to guide exploration.

Prediction error has been utilized as an exploration signal in both reinforcement learning~\cite{pathak2017curiosity,savinov2018episodic,han2020curiosity,burda2018large} and active inference~\cite{friston2012active}.
In active inference, the agent's ultimate goal is to maximize the
predictability of the outcome of environmental interactions. In contrast, reinforcement learning algorithms use a reward signal to guide agent behavior.  However, extrinsic rewards are often  sparse and specific to a given environment, so for complex problems it is useful to combine extrinsic and intrinsic rewards.

Curiosity is one form of intrinsic reward signal that has been shown to promote faster learning~\cite{pathak2017curiosity,savinov2018episodic,burda2018large} as well as generalizable behaviors that transfer well to new environments.  Pathak et al.~\shortcite{pathak2017curiosity} developed an Intrinsic Curiosity Module (ICM) which used the error in prediction of the consequences of one's own actions to promote exploration of the environment, and 
Zhelo et al.~\shortcite{zhelo2018curiosity}  applied the same module to guide exploration in mapless navigation. Han et al.~\shortcite{han2020curiosity} present a Curiosity-driven Variational Autoencoder (CVAE) that they used to guide exploration in a deep Q-learning system.  The VAE is used to predict the next state and exploration bonuses are awarded for poor  predictions.  This paper demonstrates that our proposed visual episodic memory architecture outperforms CVAE at detecting dynamic anomalies, due to its superiority at modeling temporal patterns.

\section{Method}

Figure~\ref{fig:sys_arch} illustrates our system architecture, which is divided into two components: 1) the simulation environment and 2) visual episodic memory implemented with twin Convolutional LSTM autoencoders (ConvLSTM)~\cite{convlstm}. 

In simulation, a mobile robot explores its environment providing a camera video feed to the twin convolutional recurrent autoencoders. In order to provide the mobile robot a timely exploration signal, inference latency of the deep network should be minimized and thus the two deep learning models run asynchronously. As a form of episodic visual memory, the recurrent autoencoder is trained to reconstruct 10 video frames in which the error in reconstruction is used to guide the robot to unexplored areas and dynamics.

In frontier-based exploration~\cite{horner2016}, the agent seeks to move to regions that possess a high potential for information gain, usually at the edge of the mapped area.  Rather than prioritizing these frontiers by proximity and size, our intrinsic motivation guided exploration uses sequence reconstruction error.

\begin{figure}
\begin{center}
\includegraphics[scale=.33]{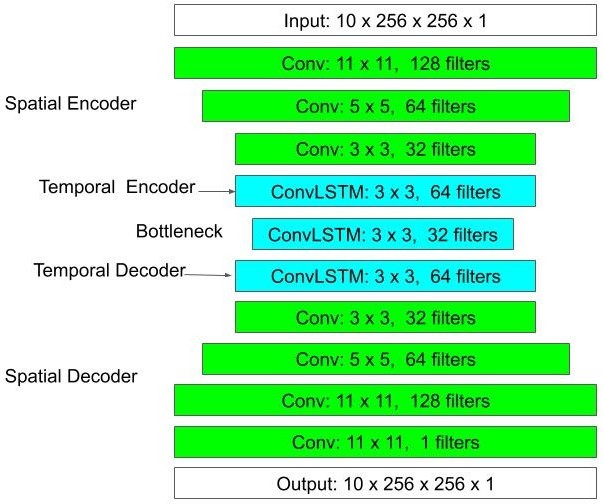}
\caption{The visual episodic memory consists of a convolutional LSTM autoencoder.  The autoencoder processes ten video frames simultaneously and attempts to reconstruct input the frames by learning the spatiotemporal patterns of the environment.  The autoencoder bottleneck forces learning of a dense representation and prevents overfitting.}
\label{fig:convLSTM}
\end{center}
\end{figure}

Figure~\ref{fig:convLSTM} shows our proposed visual episodic memory architecture which is composed of a spatial encoder (implemented with convolutional layers), temporal encoder (ConvLSTM layers), a bottleneck, and the decoder layers. ConvLSTM~\cite{convlstm} was developed for forecasting problems and has convolutions in both the input-to-state and state-to-state transitions. 


To quantify intrinsic motivation, we use Structural Similarity Index Measure (SSIM) between the ground truth view and the predicted view generated by the decoder where $\mu_x$ is the average of x, $\mu_y$ is the average of \emph{y}, $\sigma_x^2$ is the variance of \emph{x}, $\sigma_y^2$ is the variance of \emph{y}, $\sigma_{xy}$ is the covariance of \emph{x} and \emph{y}, $c_1$ and $c_2$ are small constaints included to avoid numeric instability:
\[
SSIM(\emph{x,y})=\frac{(2\mu_x\mu_y+c_1)(2\sigma_{xy}+c_2)}{(\mu^2_x\mu^2_y+c_1)(\sigma_x^2+\sigma_y^2+c_2)}
\]
Although there are many ways to calculate image differences, SSIM is better at capturing perceptually meaningful differences between images.

\begin{figure}
\begin{center}
\includegraphics[width=1.0\columnwidth]{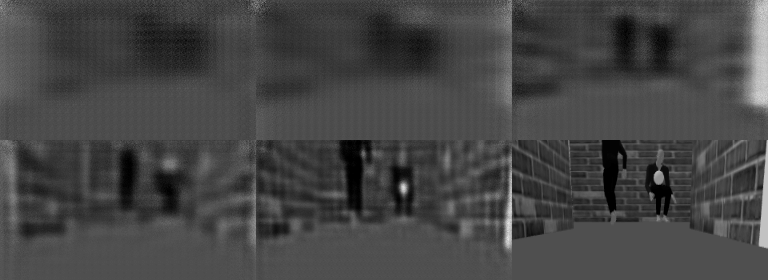}
\caption{During training the model learns to reconstruct all the sequence frames.  Starting in the upper left, five predicted frames of a dynamic scene are shown spanning 1800 epochs.  The ground truth frame is shown in the lower right.}
\label{fig:frames}
\end{center}
\end{figure}

By combining our visual episodic memory with frontier-based exploration, we ensure that our technique is robust against the ``Noisy-TV'' scenario~\cite{burda2018exploration} in which the agent is captivated by an endlessly changing visual scene. Our implementation weights both coverage and curiosity in its exploration priorities. 
Figure \ref{fig:frames} shows the progression of convolutional LSTM autoencoder learning to reconstruct a sequence of frames of moving people in the simulation.  Starting with the upper left frame, five reconstructed frames are shown with the ground truth frame on the lower right. The autoencoder learns to reconstruct the sequence of frames to include learning the dynamics of moving objects as well as that of the robot itself. 


\section{Experimental Setup}

Our experiments were conducted within the Gazebo robot simulation environment using the Turtlebot platform as the exploration agent.  The navigation stack and inter-process communication between various algorithms and the Gazebo simulator was implemented using Robot Operating System (ROS).  Agent observations consist of laser scan values, linear velocity, angular velocity, and camera output from the Turtlebot.  Of the eight rooms in the simulation, four contain static and dynamic visual anomalies while the other four are empty and featureless.

Sampling ten frames at a time from the camera, the sequence is passed to the convolutional LSTM autoencoder as inputs and as the ground truth target values for reconstruction and subsequent loss calculation such that the autoencoder learns to reconstruct the entire sequence. 
The intrinsic motivation reward provided to the frontier-based exploration is a single scalar value representing the accuracy of the sequence prediction calculated using SSIM between a ground truth image frames and the predicted frames.  

Four experimental conditions were evaluated: 1) our LSTM inference only condition using a baseline model trained on a plain environment, 2) our LSTM learning condition in which model training continues during exploration, 3) the state of the art Curiosity-driven Variational Autoencoder (CVAE)~ \cite{han2020curiosity} and 4) an unmodified frontier exploration method.  For each condition, ten trials were conducted with a time limit of 12 minutes each, which is time enough to explore two to four rooms on the map, depending on the randomly selected start location and subsequent exploration path.

Frontier exploration is implemented using \texttt{explore\_lite}~\cite{horner2016}, a ROS greedy exploration package which receives occupancy grid updates from the SLAM package.  The frontier exploration method maintains a list of frontiers and their associated cost which is calculated using both the frontier size and distance from the robot.  Our episodic memory method replaces the frontier cost with a value proportional to the novelty of the area beyond the frontier.  The inference only condition loads pre-trained weights and runs inference or forward pass at 10.5 Hz enabling fast SSIM calculations.  During the learning condition in which the ConvLSTM network continues to train during exploration, backpropagation updates are performed after each inference pass which results in an update rate of about 1.6 Hz for the learning network.

For the inference and learning conditions, both are initialized with our LSTM autoencoder model trained in an area void of anomalies. The model was trained until an SSIM value of 0.95 was achieved; a value of 1.0 indicates perfect reconstruction.

For the inference only condition, a static SSIM threshold was set; when the SSIM falls below this threshold, the robot receives a decrease in frontier cost proportional to the SSIM value to motivate exploration.  Due to the deep network inference latency, we only augment frontiers in front of the robot and take the robot's max speed into account.
This ensured that our delayed SSIM signal was produced only by the features beyond the frontiers in view.

The learning condition is nearly identical to the inference only condition except that model training was enabled, allowing the robot to learn and update the autoencoder model during exploration.  As shown in Figure~\ref{fig:sys_arch}, weights from the learning model are copied to an inference model to enable faster inference and SSIM calculation. 
The VAE curiosity method \cite{han2020curiosity} was trained and integrated into the ROS navigation stack using the same method employed by our LSTM  episodic memory.  For autoencoder network training, MSE loss is used with a learning rate of 0.0001 and the Adam optimizer. 

The experimental map (see Figure~\ref{fig:ten_SSIM}) was arranged to have eight symmetric outer rooms with every other room having anomalies.  
For each trial, the Turtlebot was started at a random location and orientation within the 4x4 center grid.



\section{Results}
\begin{figure}
\begin{center}
\includegraphics[scale=.5]{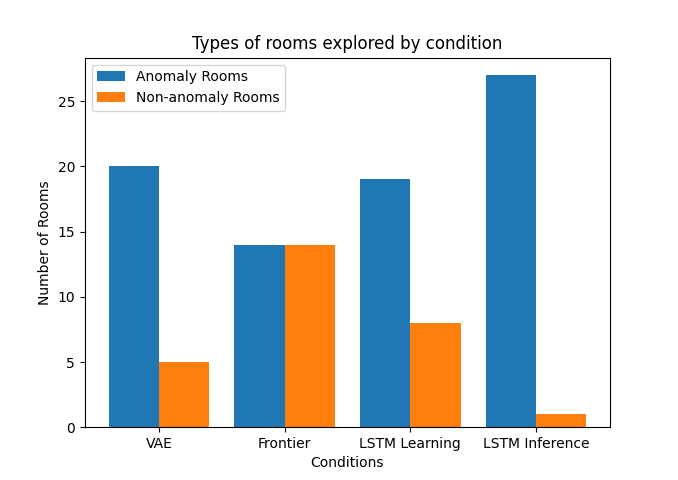}
\caption{The number of anomaly rooms (blue) vs. non-anomaly ones (orange) explored by the robot, summed over ten trials. Curiosity-driven Variational Autoencoder (CVAE) \cite{han2020curiosity} was tested with equivalent training and inference conditions.  As expected the frontier exploration method \cite{yamauchi1997frontier} is insensitive to visual anomalies and explores rooms in equal proportion. The difference between our proposed \textbf{LSTM Inference} technique and the comparison methods (\textbf{Frontier} and \textbf{VAE}) is statistically significant ($p<0.05$).}

\label{fig:roomRatio}
\end{center}
\end{figure}

Our experiments were performed on a map with both static and dynamic anomalies (moving people and objects).  We compared two versions of our proposed intrinsic motivation method (\textbf{LSTM Learning} and \textbf{LSTM Inference}) against the \textbf{VAE} based curiosity method \cite{han2020curiosity} and unmodified frontier-based exploration (\textbf{Frontier}).  Figure~\ref{fig:roomRatio} shows that the addition of the artificial episodic memory system compels the agent towards the static and dynamic anomalies, enabling the agent to consistently locate the anomalies prior to exhaustively searching other areas.  The unmodified frontier condition ignores visual changes and explores an equal number of anomalous and ordinary rooms.  The state of the art \textbf{CVAE}~\cite{han2020curiosity} prioritizes exploring visual anomalies but does not perceive subtle temporal differences as anomalies. In our proposed \textbf{LSTM Inference} condition, the episodic memory based intrinsic motivation compelled the robot to explore 27 anomaly rooms out of the 28 rooms explored, successfully searching out anomalies 96\% of the time. \textbf{LSTM Learning} encodes complex spatio-temporal patterns so if the same people and objects are encountered again (as is true in our simulation) its motivation to explore the area decreases. 


\begin{figure}
\begin{center}
\includegraphics[width=1.0\columnwidth]{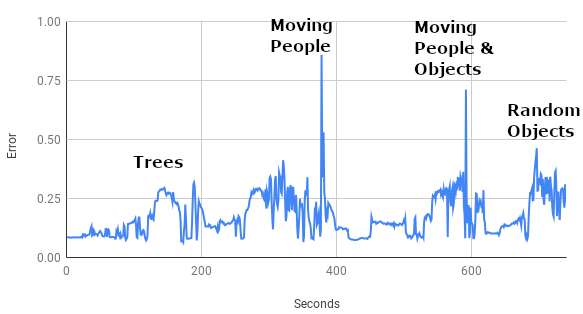}
\caption{The frame reconstruction error for the four types of anomaly rooms.  The high error spikes for moving anomalies indicate the difficulty of predicting unlearned dynamics.}
\label{fig:errorvTime}
\end{center}
\end{figure}



\begin{figure}
\begin{center}
\includegraphics[width=0.8\columnwidth]{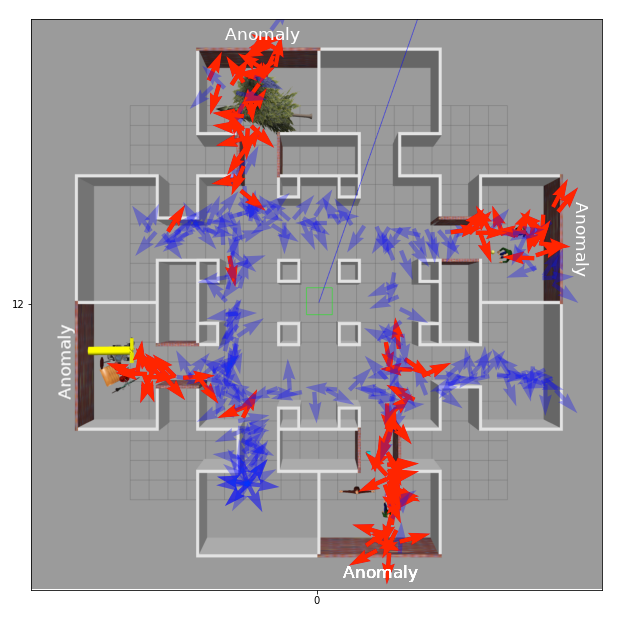}
\caption{Colored vectors indicate the SSIM values with high values as blue and low values as red.  As shown, the SSIM values are lowest in the anomaly rooms and when the robot has recently viewed into an anomaly room.}
\label{fig:ten_SSIM}
\end{center}
\end{figure}

The rest of the section presents a comprehensive analysis of our visual episodic memory architecture's ability to reconstruct common scenes.  We attempt to answer the following questions:
1) how do different types of obstacles affect reconstruction error? 2) how does robot orientation affect reconstruction error? 3) how vulnerable is our architecture to catastrophic forgetting of previous areas? 4) how successful is the LSTM at reconstructing real-world videos?
Figure \ref{fig:errorvTime} shows the SSIM value over the course of a single trial of the inference only condition.  We can see the prediction difficulty for each room as well as the areas in between rooms.  Figure~\ref{fig:ten_SSIM} shows high and low SSIM values overlaid on the map.

\begin{figure}
\begin{center}
\includegraphics[width=0.9\columnwidth]{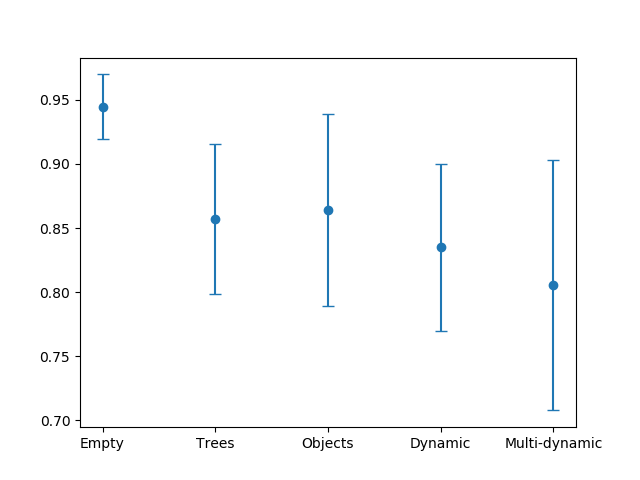}
\caption{For the inference only condition, the SSIM mean and standard deviation for each type of room indicates the relative difficulty in predicting sequences.  The episodic memory model was trained on empty rooms, generating an intrinsic motivation signal towards rooms with various types of anomalies.}
\label{fig:SSIM_uStd}
\end{center}
\end{figure}

In Figure \ref{fig:SSIM_uStd}, the SSIM mean and standard deviation for each type of room are aggregated over all trials for the inference only condition.  These data show that both the vegetation (trees) and the random static objects have similar means with the random objects having a higher standard deviation, likely due to the less consistent texturing vs. the vegetation.  The lower mean for the dynamic anomaly rooms with moving objects and people show that predicting video sequences was more difficult with the dynamic anomaly rooms having the lowest SSIM mean and the largest standard deviation.

\begin{figure}
\begin{center}
\includegraphics[width=0.9\columnwidth]{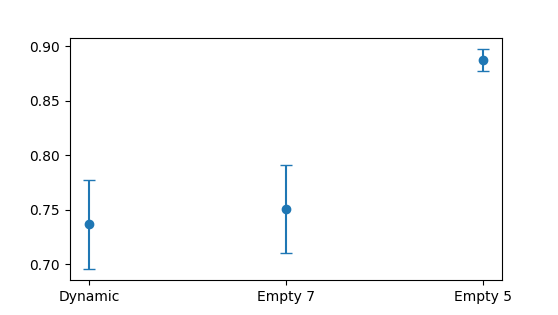}
\caption{SSIM mean and standard deviation are shown for a single learning condition trial over three rooms.  The dynamic anomaly room causes catastrophic forgetting by the model which is evident by the low mean of empty room 7.  The model quickly relearns the empty room giving a significantly higher mean for empty room 5.}
\label{fig:Order}
\end{center}
\end{figure}

In the learning condition, we observe that when new spatiotemporal features are learned, previous features are sometimes forgotten, a behavior known as catastrophic forgetting~\cite{kirkpatrick2017overcoming,toneva2018empirical}.  Figure \ref{fig:Order} shows three rooms explored during a trial in exploration order.  As expected, the initial dynamic anomaly room had a low mean SSIM of about 0.73.  However, the learning that occurred during exploration of the anomaly room caused the model to forget some of the baseline features.  Specifically, the next room explored was empty and had a mean SSIM value of 0.75, well below the 0.94 for the inference only condition (shown in Figure \ref{fig:SSIM_uStd}).  Consequently, learning in the empty room enables the autoencoder to quickly relearn features and more accurately reconstruct frames in the next empty room explored, as shown by the mean SSIM of 0.89, still below the inference only condition.  



\begin{figure}
\begin{center}
\includegraphics[width=1.1\columnwidth]{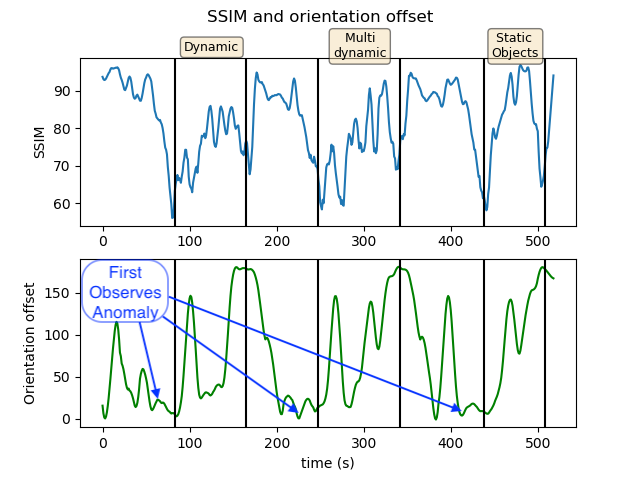}
\caption{The upper blue graph shows SSIM values for a single inference trial, and the lower green graph shows relative orientation over three rooms, indicating when the robot first observes the anomaly.  The three rooms explored were a dynamic anomaly, a multi-dynamic anomaly and static objects room.}  
\label{fig:infSSIMvTime}
\end{center}
\end{figure}

To illustrate the relationship between the SSIM value and robot orientation relative to the anomaly, Figure~\ref{fig:infSSIMvTime} plots SSIM and relative orientation values over three rooms.  The lower plot is the orientation offset relative to the direction of the anomaly, showing when the robot first observes the anomaly.  In the first dynamic anomaly room, the SSIM value decreases as the robot looks into the room.  Once inside the room, the robot rotates to explore the area where the moving people are intermittently obstructing the view, causing the SSIM value to fluctuate. Lastly, upon exiting the room, the SSIM value increases again until the next dynamic anomaly room is visible where we observe another steady reduction in SSIM value.

\begin{figure}
\begin{center}
\includegraphics[scale=.5]{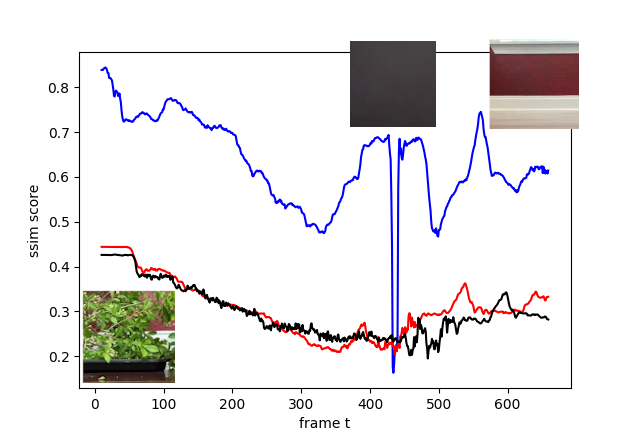}
\caption{SSIM scores for real world video collected from a camera mounted on a small autonomous mobile robot.  SSIM values are shown for the base case (blue), vegetation (red) and random motion (black).  Note, the large dip in the base case SSIM score is the result of bumping an obstacle, momentarily blocking out the camera lens.}
\label{fig:Real_wold}
\end{center}
\end{figure}

Figure~\ref{fig:Real_wold} shows the performance of our visual episodic memory at reconstructing  real world video from a camera mounted on a small autonomous mobile robot. SSIM values are shown for the base case (blue), vegetation (red) and moving vegetation (black). As with the simulation environment, the robot explored and learned the non-anomaly environment until the SSIM score for sequence reconstruction reached 0.95. Then, a vegetation anomaly was inserted into the environment for testing.  Note, the large dip in the base case SSIM score is the result of bumping an obstacle, momentarily blocking out the camera lens. Interestingly, the static and dynamic anomalies have comparable SSIMs.   Examining the dynamic anomaly video, in cases where the objects move too quickly and reveal much of the familiar pre-trained background, both static and dynamic obstacles will exhibit similar SSIM scores.

\begin{figure}
\begin{center}
\includegraphics[width=0.8\columnwidth]{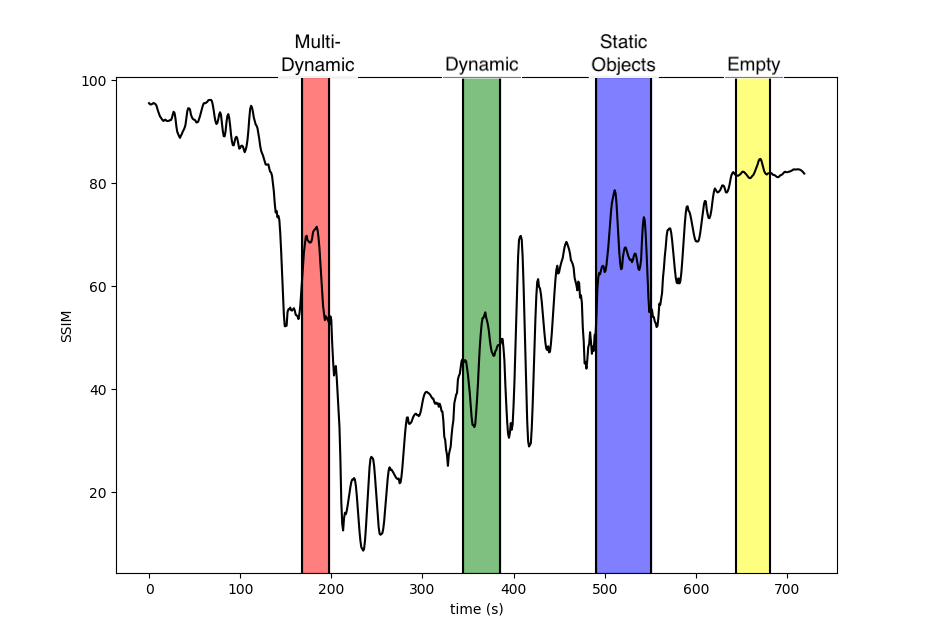}
\caption{When observing the SSIM value for a learning condition trial over four rooms, the initial catastrophic forgetting is evident after the first room.  Spatiotemporal feature relearning happens quickly thereafter.}
\label{fig:Forget}
\end{center}
\end{figure}

\section{Conclusion and Future Work}


Our visual episodic memory successfully reconstructs real-world videos and is well suited for robotic applications such as search and rescue (SAR), site security, and other applications such as casualty rescue robots where finding visual anomalies is crucial.  From the results we can see that the SSIM-based intrinsic motivation greatly helped the robot in its ability to explore anomaly rooms within the shortest time, consistently seeking out both static and dynamic anomalies in the environment.  The inference only version outperforms Curiosity-driven Variational Autoencoder (CVAE) at driving exploration since the ConvLSTM is superior at encoding spatio-temporal patterns and detecting the dynamic anomalies that are missed by the VAE. The learning condition in which the model is continuously trained during exploration, exhibited signs of catastrophic forgetting which could be mitigated with selective adjustments to the learning rate and weight plasticity.


Our future work will focus on full environmental interaction such that the agent seeks out new interactions thus exploring not only the environment but also its own range of interaction capabilities across various dynamic environmental conditions.  We also plan to explore the usage of the sequence frame structural similarity index measure as an intrinsic reward signal for reinforcement learning systems.



\bibliographystyle{flairs}
\bibliography{convLSTM_RL}

\end{document}